\documentclass[]{bytedance_seed}

\usepackage[toc,page,header]{appendix}

\usepackage{minitoc}
\usepackage{hyperref}
\usepackage{url}
\usepackage{graphicx}
\usepackage{booktabs}
\usepackage{multirow}
\usepackage{colortbl}
\usepackage{xcolor}  % 加载扩展颜色集
\usepackage{enumitem}
\usepackage{amsmath,amssymb}
\usepackage{tikz}
\usepackage{caption}
\usepackage{subcaption}
\usepackage{makecell}
\usepackage{wrapfig}

\usepackage{hyperref} 
\usepackage{bm}
\usepackage{float}
\usepackage{multirow}
\usepackage{xcolor}
\usepackage{makecell}
\usepackage{booktabs}
\usepackage{amsmath}
\usepackage{graphicx}

\definecolor{citeblue}{rgb}{0,0,0.8} 
\let\oldcite\cite % 先保存原始\cite命令

\usepackage[table]{xcolor} % 支持 \rowcolor
\usepackage{booktabs}      % 支持 \midrule
\usepackage{multirow}      % 支持 \multirow

\renewcommand{\cite}[1]{{\color{citeblue}\oldcite{#1}}} % 包裹颜色命令，不影响其他格式

\definecolor{Gray}{gray}{0.94}

\usepackage{listings}
\lstdefinestyle{promptStyle}{
    language=Python, % Using Python style for basic syntx highlighting, adjust as needed
    basicstyle=\small\ttfamily,
    backgroundcolor=\color{gray!10},
    frame=single,
    framesep=10pt,
    breaklines=true,
    columns=fullflexible,
    showstringspaces=false,
    commentstyle=\color{blue}, % For comments
    keywordstyle=\color{purple}, % For keywords
    stringstyle=\color{green!60!black}, % For strings
}
\definecolor{aliceblue}{rgb}{0.94, 0.97, 1.0}

\title{DreamActor-M2: Universal Character Image Animation via Spatiotemporal In-Context Learning}

\author[123,*]{Mingshuang Luo}
\author[1,*]{Shuang Liang}
\author[1,*]{Zhengkun Rong}
\author[1,\dagger]{Yuxuan Luo}
\author[1,\S]{Tianshu Hu}
\author[2,\S]{Ruibing Hou}
\author[23]{Hong Chang}
\author[4]{Yong Li}
\author[1]{Yuan Zhang}
\author[1]{Mingyuan Gao}

\affiliation[1]{ByteDance Intelligent Creation \\}
\affiliation[2]{Key Lab of Intell. Info. Process., ICT, CAS \\}
\affiliation[3]{University of Chinese Academy of Sciences \\}
\affiliation[4]{Southeast University}

\contribution[*]{Equal Contribution}
\contribution[\dagger]{Project Lead}
\contribution[\S]{Corresponding Authors}

\abstract{
Character image animation aims to synthesize high-fidelity videos by transferring motion from a driving sequence to a static reference image. 
Despite recent advancements, existing methods suffer from two fundamental challenges: 
(1) suboptimal motion injection strategies that lead to a trade-off between identity preservation and motion consistency, manifesting as a "see-saw", and
(2) an over-reliance on explicit pose priors (e.g., skeletons), which inadequately capture intricate dynamics and hinder generalization to arbitrary, non-humanoid characters.
To address these challenges, we present \textbf{DreamActor-M2}, a universal animation framework that reimagines motion conditioning as an \textit{in-context} learning problem. 
Our approach follows a two-stage paradigm. First, we bridge the input modality gap by fusing reference appearance and motion cues into a unified latent space, enabling the model to jointly reason about spatial identity and  temporal dynamics by leveraging the generative prior of foundational models. 
Second, we introduce a self-bootstrapped data synthesis pipeline that curates pseudo cross-identity training pairs, facilitating a seamless transition from pose-dependent control to direct, end-to-end RGB-driven animation. 
This strategy significantly enhances generalization across diverse characters and motion scenarios.
To facilitate comprehensive evaluation, we further introduce \texttt{\textit{AW}Bench}, a versatile benchmark encompassing a wide spectrum  of characters types and motion scenarios.
Extensive experiments demonstrate that DreamActor-M2 achieves state-of-the-art performance, delivering superior visual fidelity and robust cross-domain generalization.
}

\date{\today}
\correspondence{\email{tianshu.hu@bytedance.com}, \email{houruibing@ict.ac.cn}}

\checkdata[Project Page]{\url{https://grisoon.github.io/DreamActor-M2/}}

\begin{document}
\maketitle

%不需要目录就注释掉 注意目录不要和第一页放在一块 要有\newpage
%\newpage
%\tableofcontents
%\newpage

\begin{figure}[h!]
    \centering
    % \vspace{-15pt}
    \includegraphics[width=\columnwidth]{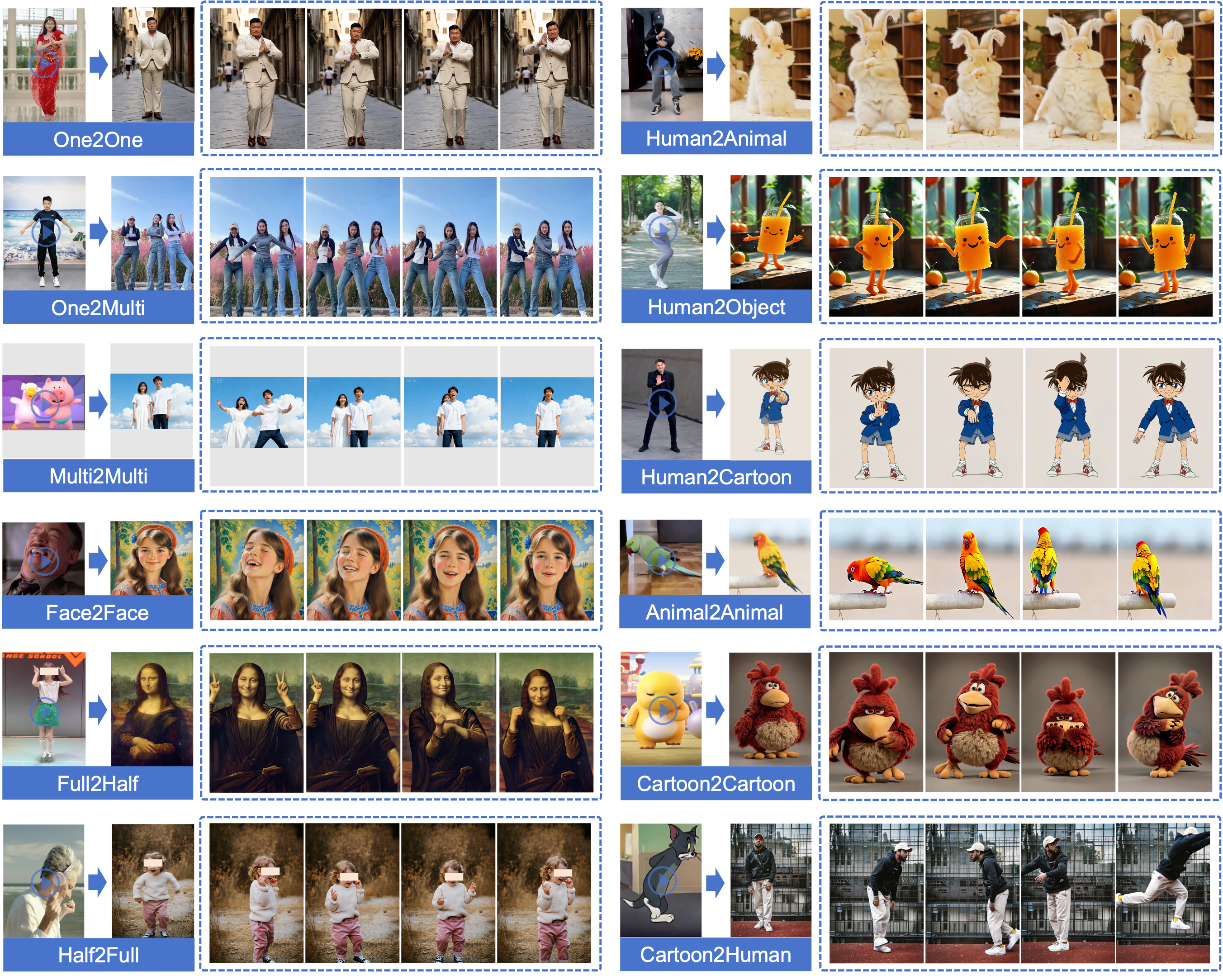}
    % \vspace{-8pt}
       \caption{The proposed \textbf{DreamActor-M2} exhibits strong generalization capability, producing diverse animations while preserving consistent character appearance. The arrow $\xrightarrow{}$ denotes the transfer of character motion from the driving video to the character depicted in the reference image.}
    \label{fig:intro}
\end{figure}
\section{Introduction}
Character image animation~\cite{hu2023animate,wang2025unianimate,tan2024animate,tan2025animate,luo2025dreamactor,gan2025humandit,zhang2024mimicmotion,cheng2025wan,chang2023magicpose} aims to synthesize high-fidelity videos by transferring motion from a driving sequence to a static reference image, a task with vast potential in digital entertainment.
While recent foundational video diffusion models~\cite{blattmann2023stable,yang2024cogvideox,wan2025wan,seawead2025seaweed,gao2025seedance,seedance2025seedance15pronative} provide powerful generative priors, effectively adapting them for animation without compromising their intrinsic generative capabilities remains a challenge.
Moreover, concurrently achieving robust identity preservation,accurate motion fidelity, and strong cross-domain generalization still pose a major challenge for existing frameworks.

These limitations primarily stem from two fundamental drawbacks. First, prevailing motion injection strategies exhibit an inherent trade-off between identity preservation and motion consistency. Specifically, pose-aligned channel-wise injection methods~\cite{hu2023animate,wang2025unianimate,zhang2024mimicmotion,lidispose} frequently suffer from "shape leakage", whereby the structural priors embedded in the driving signal distort the reference identity.  Conversely, cross-attention based methods~\cite{ding2025mtvcrafter,tan2024animate,tan2025animate} often overly compress motion representations, resulting in the loss of fine-grained temporal dynamics and degraded motion coherence.  This "see-saw" effect prevents existing frameworks from simultaneously achieving high-fidelity animation and stable identity preservation.
Second, the heavy reliance on explicit pose priors (e.g., skeletons) imposes a representation bottleneck that restricts model flexibility and generalization~\cite{luo2025dreamactor,zhu2024champ,cheng2025wan,yan2025scail}. Such pose estimators~\cite{yang2023dwpose,SMPL:2015} are inherently error-prone in complex human-centric dynamics and, more critically, are intrinsically  incapable of generalizing to arbitrary non-humanoid characters, including cartoons and animals. Although implicit motion representations have been explored to mitigate this dependency~\cite{zhang2025flexiact, song2025x}, these methods either remain tethered to pose-derived supervision during training or necessitate costly per-video fine-tuning to capture motion-specific details. Consequently, such constraints substantially impede their scalability and applicability in diverse real-world scenarios.

To address these challenges, we propose \textbf{DreamActor-M2}, a universal character animation framework that reframes motion conditioning as an \textit{in-context learning} problem. Departing from traditional approaches that rely on complex motion injection modules, our method adopts a simple yet effective design: motion control signals are spatiotemporally concatenated with the reference image to construct a unified input representation. Through this design, the pre-trained video backbone can naturally interpret motion cues as visual context, thereby effectively bridging the modality gap between appearance and motion. This formulation preserves the original model architecture while fully exploiting the inherent generative priors of foundation video models, enabling high-fidelity animation without compromising their intrinsic capabilities. 

Our framework evolves through a strategic two-stage paradigm. In the first stage, we develop \textit{Pose-based DreamActor-M2}, which leverages augmented 2D skeletons as an initial form of motion context. To alleviate the semantic insufficiency of pose-based conditioning, we incorporate a  target-oriented motion-semantic guidance module powered by Multimodal Large Language Models (MLLMs). This module aligns visual cues with fine-grained semantic descriptions, enabling the model to simultaneously preserve identity fidelity and motion consistency. 
In the second stage, we advance to \textit{End-to-End DreamActor-M2}, with the goal of removing the reliance on external pose estimators. To this end, we propose a self-bootstrapped data synthesis pipeline that exploits the pose-based variant to curate a high-quality pseudo-paired dataset. Training on these synthesized cross-identity pairs allows the end-to-end model to derive motion directly from raw RGB sequences, without any explicit pose supervision. This progressive transition not only circumvents the inherent limitations of pose estimation but also substantially extrapolates the model's generalization capability to arbitrary characters and complex motion scenarios.

To facilitate a rigorous evaluation, we introduce \texttt{\textit{AW}Bench}, a comprehensive benchmark encompassing a wide spectrum of character categories and motion types. Extensive experimental results demonstrate that DreamActor-M2 achieves state-of-the-art performance in term of visual fidelity and cross-domain generalization.

Our key contributions are summarized as follows:
\begin{itemize}
\item We propose a spatiotemporal in-context motion conditioning strategy that bridges modality gaps and effectively balances identity preservation with motion consistency for character image animation.
\item We propose a self-bootstrapped synthesis-and-training pipeline that enables a seamless transition to end-to-end, pose-free animation, substantially improving generalization across diverse character domains.
\item We construct a comprehensive benchmark encompassing diverse motion patterns and character categories, offering the community a more challenging and rigorous evaluation platform for character image animation.
\end{itemize}
\section{Related Work}
\textbf{Latent Video Diffusion Models.} Diffusion-based generative models~\cite{blattmann2023stable,yang2024cogvideox,wan2025wan,kong2024hunyuanvideo,seawead2025seaweed,gao2025seedance,zhang2025waver} have recently achieved remarkable success in video generation. Notably, Wan2.1~\cite{wan2025wan} and Seedance 1.0~\cite{gao2025seedance} have emerged as high-performance video foundation models, supporting both text-to-video and image-to-video generation. Since character image animation inherently aligns with the image-to-video setting, we adopt Seedance 1.0 as the pre-trained backbone for our DreamActor-M2 framework.

\noindent \textbf{Pose Guidance in Character Image Animation.} Pose-guided approaches have emerged as the dominant paradigm for character image animation.
A line of works~\cite{hu2023animate,zhu2024champ,xu2024magicanimate,zhang2024mimicmotion,wang2025unianimate,gan2025humandit,chang2025x,he2025posegen,tu2025motionfollower} inject 2D skeleton or SMPL signals as motion conditions via \textit{channel-wise} concatenation or additive fusion. While ensuring spatial alignment and motion consistency, training on same-identity data often induces \textit{identity leakage}, where identity-specific appearance cues become entangled with motion features, substantially degrading cross-identity generalization. 
To address this issue, alternative methods ~\cite{ding2025mtvcrafter} adopt cross-attention injection mechanisms. In these approaches, an auxiliary pose encoder compresses motion signals into latent representations, which are then injected into the generation backbone through cross-attention. This strategy often decouples pose from identity by employing skeleton scaling~\cite{AnimateX2025} or SMPL normalization. Additionally, \cite{kim2025temporal} concatenates conditioning and target frames along the temporal dimension, leveraging the backbone's temporal modeling capacity for capturing global motion patterns. However, such designs inevitably distort fine-grained motion semantics, ultimately compromising motion fidelity.

Another line of research~\cite{zhang2025flexiact,wei2024dreamvideo} avoids explicit motion representations altogether, instead directly exploiting raw RGB frames for motion guidance. DreamVideo~\cite{wei2024dreamvideo} adopts a "one-model-per-identity" paradigm, which severely limits generalization and necessitates retraining for each novel character. 
 FlexiAct~\cite{zhang2025flexiact} requires a dedicated frequency-aware embedding for each driving video, introducing substantial computational overhead.
X-Unimotion~\cite{song2025x} leverages implicit motion representations during generation, yet still relies on 2D pose signals as supervision during training, and thus remains constrained by inherent limitations of pose-based methods.

\begin{figure}[t]
    \centering
    \includegraphics[width=\columnwidth]{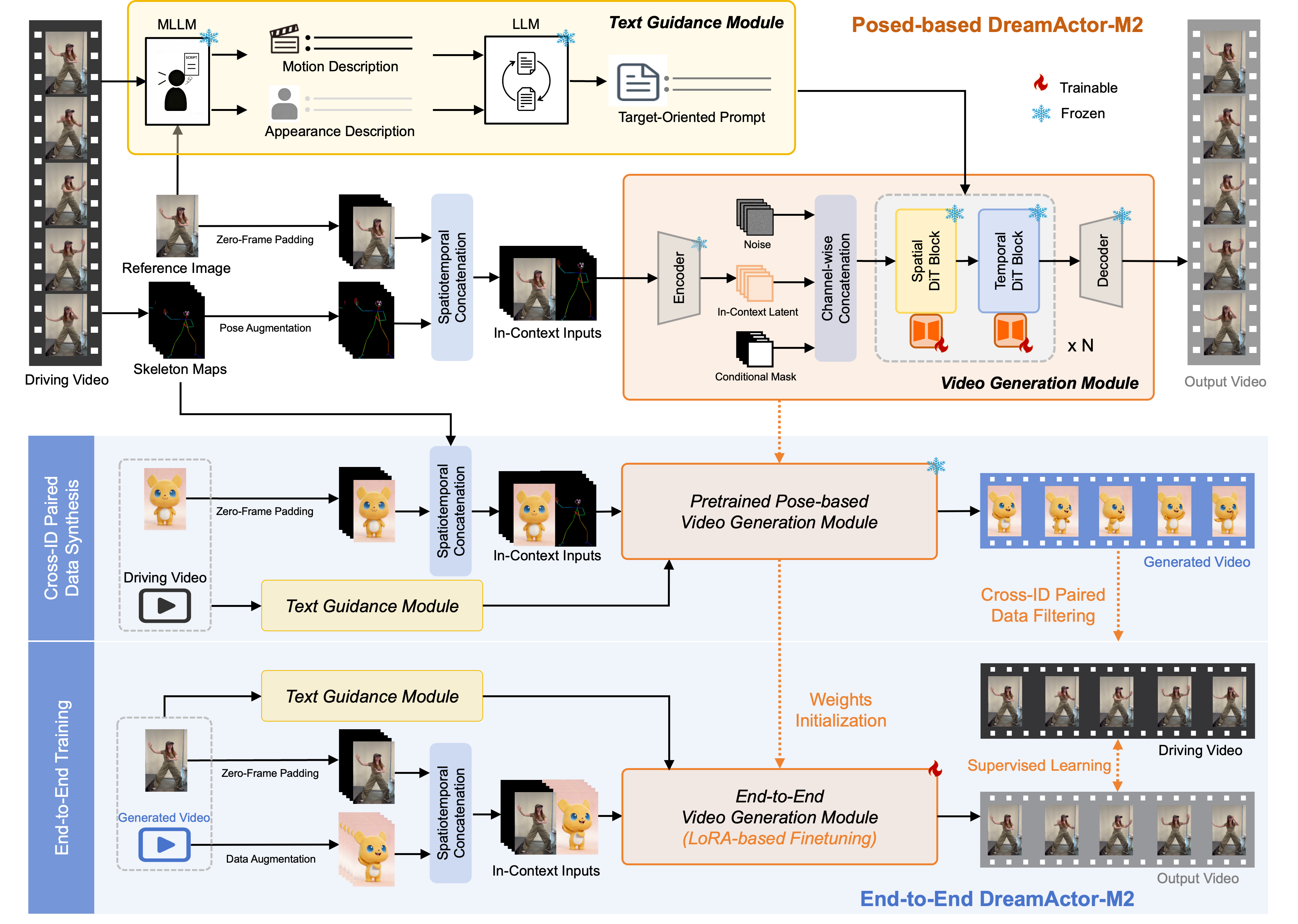}
    % \vspace{-0.2cm}
    \caption{The schematic overview of proposed DreamActor-M2.}
    \label{fig:pipeline}
    % \vspace{-0.2cm}
    \label{overview}
\end{figure}

\noindent \textbf{In-Context Learning.}
Despite its success in large language models (LLMs)~\cite{brown2020language,dong2022survey,rubin2021learning,highmore2024context}, vision-language models (VLMs) ~\cite{doveh2024towards,alayrac2022flamingo,nulli2024context}, and image generation~\cite{huang2024group, huang2024context, sun2024x,li2025visualcloze,labs2025flux}, in-context learning (ICL) remains relatively underexplored in video generation. TIC-FT~\cite{kim2025temporal} temporally concatenates condition and target frames to boost fidelity and training/generation efficiency, it does not address the unique challenges of character animation. 
A concurrent work, SCAIL~\cite{yan2025scail},leverages full 3D pose sequences as contextual inputs to achieve studio-grade animation quality. However, its performance is highly contingent on accurate 3D motion representations, which limits its applicability in practical settings where such structural signals are noisy, incomplete, or unavailable.
In contrast, our end-to-end DreamActor-M2 eliminates explicit pose dependencies by directly interpreting raw motion signals as contextual inputs. 
This design significantly enhances generalization across diverse character types and unconstrained motion patterns.
\section{Approach}
This section presents DreamActor-M2, a universal character animation framework conditioned on a reference image and driving signals (e.g., pose sequences or video clips). The methodology is organized as follows: Sec. \ref{sec3.1} briefly reviews the Diffusion Transformer (DiT) backbone that serves as the foundation of our approach. Sec. \ref{sec3.2} introduces our core in-context motion injection strategy. Building upon this mechanism, Sec. \ref{sec3.3} elaborates on the Pose-based DreamActor-M2 framework. Finally, Sec. \ref{sec3.4} details the evolution toward an End-to-End DreamActor-M2 pipeline that eliminates the reliance on explicit pose priors. A schematic overview of the proposed framework is provided in Fig.~\ref{overview}. 

\subsection{Preliminary}
\label{sec3.1}
\textbf{Latent Diffusion Model.}
Our framework is built on the Latent Diffusion Model (LDM), where a Variational Autoencoder (VAE) encodes input images $I$ into latent representations $\mathbf{z}=\xi(I)$. Gaussian noise $\epsilon$ is progressively injected into latents $\mathbf{z}_t$ at different timesteps, with the optimization objective:
\begin{equation}
\mathcal{L} = \mathbb{E}_{\mathbf{z}_t,c,\epsilon,t}\left(\left\|\epsilon-\epsilon_\theta\left(\mathbf{z}_t,c,t\right)\right\|_2^2\right)
\end{equation}
where $\epsilon_\theta$ denotes the denoising network and $c$ is the conditional input. At inference, noise latents are iteratively denoised and reconstructed into images via VAE decoder. 
We adopt Seedance 1.0 \cite{gao2025seedance} as our backbone, which employs the MMDiT architecture \cite{esser2024scaling} to support multi-modal and multi-task video generation.

\subsection{Motion Injection via In-Context Learning}
\label{sec3.2}
Existing motion injection methods suffer from an inherent trade-off between identity preservation and motion consistency. To contextualize our approach, we first review representative alternatives before introducing our spatiotemporal in-context injection strategy.

\textit{Alternative 1: Pose Alignment Injection.}
Several works~\cite{hu2023animate,wang2025unianimate} encode 2D pose signals into latent representations and inject them into noise latents via channel-wise concatenation or additive fusion. Such designs enforce motion consistency through strict spatial alignment, however, pose encoding inevitably carry pose shape and structural cues, leading to dentity leakage and noticeable identity distortion during animation. 

\textit{Alternative 2: Cross-Attention Injection.}
Other methods~\cite{ding2025mtvcrafter} compress motion signals into latents via an auxiliary pose encoder and inject them into the generation backbone via cross-attention. By decoupling motion control from explicit spatial alignment, these methods partially alleviate identity leakage. Nevertheless, the required latent compression often discards fine-grained motion details, frequently resulting in temporally or anatomically unnatural animations.

\textit{Alternative 3: Temporal-Level In-Context Injection.}
As explored in~\cite{kim2025temporal}, conditioning and target frames can be concatenated along the temporal dimension, allowing the backbone's temporal modeling capacity to learn global motion patterns effectively. However, the absence of frame-wise spatial correspondence tends to degrade fine-grained motion details and  leads to suboptimal reconstruction quality.

\textit{Our Method: Spatiotemporal In-Context Injection.} Inspired by in-context learning (ICL) in LLMs~\cite{brown2020language, highmore2024context} and VLMs~\cite{alayrac2022flamingo, doveh2024towards, nulli2024context}, which achieves seamless task adaptation via direct input integration, we propose a spatiotemporal ICL strategy to resolve the aforementioned trade-off. 
To model the video's spatiotemporal structure, we create a unified sequence by: (1) spatially concatenating the reference image and the first motion frame as a hybrid anchor, (2) aligning the subsequent motion frames with a reference-sized blank mask, and (3) temporally stacking all frames. This approach avoids lossy compression, bridges modality gaps, and unleashes the pre-trained model's potential for superior identity and motion fidelity.

\textbf{In-Context Operation.}
Our objective is to generate a video where the subject identity of $I_\mathrm{ref} \in \mathbb{R}^{H\times W \times 3}$ follows driving motion signals $\mathbf{D} \in \mathbb{R}^{T\times H\times W \times 3}$. We construct a composite input sequence $\mathbf{C} \in \mathbb{R}^{T\times H\times 2W\times3}$ as follows:
\begin{equation}
\mathbf{C}\left[t\right] = 
\begin{cases}
I_{\mathrm{ref}} \oplus \mathbf{D}\left[t\right], & t=0, \\
\mathbf{0} \oplus \mathbf{D}\left[t\right], & t>0.
\end{cases}
\end{equation}
where $\oplus$ denotes spatial concatenation along the width axis, and $\mathbf{0}$ is a zero image matching $I_\mathrm{ref}$ dimensions. To guide the model, we construct a motion mask $\mathbf{M}_{m}$ (all elements set to 1) for motion region highlighting, and a reference mask $\mathbf{M}_r$ (1 for the first frame, 0 otherwise) to distinguish the identity source. Their spatial concatenation yields $\mathbf{M}=\mathbf{M}_r  \oplus \mathbf{M}_m$. The composite video $\mathbf{C}$ is projected into the latent sequence $\mathbf{Z}$ via a 3D VAE. Finally, $\mathbf{Z}$, noise latent $\mathbf{Z}_{\mathrm{noise}}$, and mask $\mathbf{M}$ are channel-concatenated to serve as the comprehensive input for the diffusion transformer.

\subsection{Pose-Based DreamActor-M2}
\label{sec3.3}
Pose-Based DreamActor-M2 utilizes 2D pose skeletons as motion signals under a self-supervised training paradigm. Given a video $\mathbf{V}$, we extract a pose sequence $\mathbf{P}$ as the driving signal and designate the first frame $I_\mathrm{ref}=\mathbf{V}\left[0\right]$ as the reference image. The model is then trained to reconstruct the original video $\mathbf{V}$.

\textbf{Pose Augmentation.}
2D skeletons inherently encode "body shape" cues (e.g., limb length and body proportion), which can induce identity leakage and degrade cross-identity generalization. 
To mitigate this issue while preserving the underlying motion dynamics, we apply two complementary pose augmentation strategies:
(1) \textit{Random Bone Length Scaling:} Skeleton bones are grouped into anatomical segments and subjected to random scaling. This operation perturbs limb proportions while maintaining joint connectivity and temporal motion patterns, thereby decoupling structural priors from motion dynamics.
(2) \textit{Bounding-Box-Based Normalization:} Joint coordinates are normalized with respect to the bounding box enclosing all joints in a clip. This normalization removes absolute spatial dependencies and produces a scale-invariant pose representation.

\textbf{Text Guidance.}
While pose augmentation alleviates identity leakage, it may inadvertently weaken fine-grained motion semantics. For example, perturbations can obscure subtle pose configurations such as clasped hands in a “prayer” gesture. To compensate for this loss, we introduce a target-oriented text guidance mechanism that explicitly integrates motion semantics with appearance descriptions. 
Specifically, a multimodal large language model (MLLM) is first employed to parse the driving video $\mathbf{V}$ into motion semantics $T_{\text{m}}$ (e.g., "a person is waving both hands") and to analyze the reference image $I_\mathrm{ref}$ to obtain appearance semantics $T_{\text{a}}$ (e.g., "a gray bird with colorful feathers").
An LLM subsequently fuses these descriptions into a target-oriented prompt $T_{\text{fusion}}$ (e.g., "a gray bird with colorful feathers, is waving its wings").
This fused prompt serves as a high-level semantic prior that complements low-level pose signals, improving motion controllability and enhancing the expressiveness of the synthesized animation.

\textbf{LoRA Fine-tuning.}
The pre-trained diffusion backbone inherently encodes strong generative priors for structural coherence and temporal consistency. By treating in-context images and textural descriptions as native input modalities, our framework enables seamless integration without architectural modifications, thereby preserving the model's intrinsic generative capacity.
For efficient adaptation, we adopt lightweight LoRA fine-tuning, freezing the backbone parameters and inserting LoRA modules exclusively into feed-forward layers. Notably, the text branch is excluded from adaptation to maintain robust semantic alignment. This design achieves plug-and-play customization with minimal computational overhead.

\subsection{End-to-End Training Paradigm}
\label{sec3.4}
Pose-based DreamActor-M2 relies on explicit pose estimation, which constrains its applicability in complex or non-human animation scenarios.
To overcome this, we introduce an end-to-end variant capable of processing raw RGB frames as motion signals, denoted as \textbf{End-to-End DreamActor-M2}.
The training procedure involves two stages: (i) data synthesis and quality filtering; (ii) model optimization.

\textbf{Data Synthesis and Quality Filtering.}
A key obstacle in end-to-end training lies in the absence of large-scale paired data that simultaneously exhibit motion consistency and cross-identity diversity. To overcome this, we propose a \textit{self-bootstrapped data synthesis pipeline} that leverages the pre-trained Pose-based DreamActor-M2 to generate high-fidelity pseudo-paired supervision.
Formally, given a driving video $\mathbf{V}_{\mathrm{src}}$, we extract its pose sequence $\mathbf{P}_{\mathrm{src}}$ as motion signals. Combined with a reference image $I_o$, $\mathbf{P}_{\mathrm{src}}$ is fed to the pose-based  DreamActor-M2 $\mathcal{M}_{\mathrm{pose}}$ to synthesize a new video $\mathbf{V}_o$:
\begin{equation}
\mathbf{V}_o=\mathcal{M}_{\mathrm{pose}}\left(\mathbf{P}_{\mathrm{src}}, I_o\right).
\end{equation}
The synthesized video $\mathbf{V}_o$ preserves the motion dynamics of $\mathbf{V}_{\mathrm{src}}$ while adopting a novel subject identity, thereby forming a pseudo-pair sample $\left(\mathbf{V}_{\mathrm{src}}, \mathbf{V}_o\right)$. 

To ensure the reliability of the generated supervision, we employ a dual-stage quality filtering strategy. Specifically, we first perform automatic scoring using Video-Bench~\cite{han2025video}, followed by manual verification focusing on identity fidelity and motion coherence. Only high-quality, semantically consistent pairs are retained for subsequent end-to-end training.

\textbf{Model Optimization.}
For end-to-end training, we treat $\mathbf{V}_o$ as driving video and $I_{\mathrm{ref}}=\mathbf{V}_{\mathrm{src}}\left[0\right]$ as reference image, yields a dataset:
\begin{equation}
\mathcal{D} = \left\{\left(\mathbf{V}_o, I_{\mathrm{ref}}, \mathbf{V}_{\mathrm{src}}\right)\right\}.
\end{equation}
Each triplet supervises the model to reconstruct $\mathbf{V}_{\mathrm{src}}$ from $(\mathbf{V}_o, I_{\mathrm{ref}})$, thereby learning to transfer motion patterns directly from raw RGB sequences.
To facilitate stable and efficient optimization, we warm-start with pre-trained Pose-based DreamActor-M2 model.
This initialization accelerates convergence and allows the model to inherit robust motion priors learned from explicit pose supervision, substantially improving training efficiency.
 
By learning motion transfer directly from raw RGB inputs, this paradigm bypasses the need for intermediate pose representations. Together, the Pose-based and End-to-End variants form a unified and versatile framework for character animation. To the best of our knowledge, this work presents the first fully end-to-end solution in this domain, making a significant step toward scalable and practical character animation systems.
\section{\texttt{\textit{AW}Bench}}
The core objective of DreamActor-M2 is to enable universal character image animation: it takes a driving videos of any subject (which can be human or non-human) as input, and generates a animated video based on a reference image of any subject (also human or non-human).
However, existing evaluation datasets or their setups fail to meet the requirements for evaluating our framework. For instance, some datasets~\citep{ jafarian2021learning, zablotskaia2019dwnet} are only applicable to human animation tasks and cannot support broader subject scopes. While other datasets~\citep{tan2024animate, tan2025animate} have expanded the subjects of reference images from humans to anthropomorphic characters, their coverage remains insufficient and lacks universality. In practical scenarios, our goal is to use driving videos containing one or more arbitrary subject types (e.g., humans, animals, cartoon characters, etc.) to animate reference images that feature one or more arbitrary subjects (e.g., humans, animals, cartoon characters, etc.).

\begin{figure}[t]
    \centering
    \includegraphics[width=1.0\columnwidth]{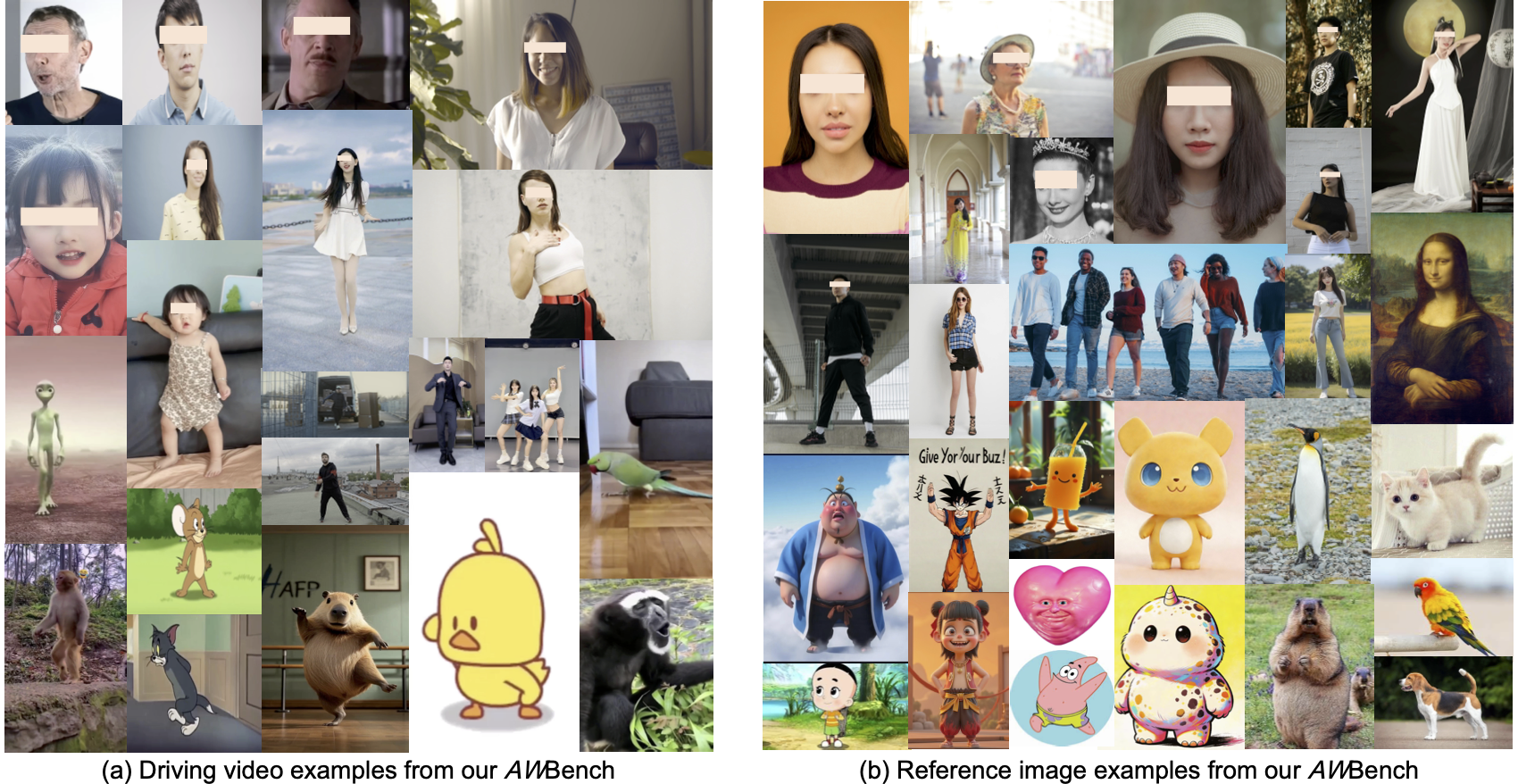}
    % \vspace{-15pt}
    \caption{Visual examples of driving video corpus and reference image corpus.}
    \label{fig:awbench}
\end{figure}

To comprehensively evaluate the efficacy and generalizability of our DreamActor-M2, we propose the \textit{"\textbf{A}nimate in the \textbf{W}ild"} \textbf{Bench}mark (\texttt{\textit{AW}Bench}), which encompasses a wide range of motion types and reference identities. The benchmark consists of 100 driving videos and 200 reference images, where the driving corpus covers human as well as non-human motion categories. 
Human motions are sampled across different body regions (face, upper body, full body), age groups (child, young adult, elderly), and activity categories (e.g., dancing, daily activities), and include both camera-tracked and static-camera sequences. Non-human motions include videos of animals (such as cats, chickens, parrots, monkeys, and orangutans) and animated characters (such as Tom the cat, Jerry the mouse, groundhogs, and cartoon aliens). 
The reference image corpus includes subjects as rich and diverse in types as those in the driving video corpus.
We also explore multi-subject driving scenarios that have not been investigated in existing works, including many-to-many and one-to-many driving. Consequently, our \texttt{\textit{AW}Bench} further includes multi-subject motion videos and multi-subject reference images. 
Finally, after data collecting and filtering, \texttt{\textit{AW}Bench} contains 100 driving videos and 200 reference characters. 
Visual examples of the driving video corpus and reference image corpus are shown in Fig.~\ref{fig:awbench}.
\section{Experiments}
\subsection{Implementations}
\textbf{Implementation Details.}
To train the pose-based DreamActor-M2, we employ two skeleton augmentation strategies: bone length scaling ($\mathcal{U}(0.8, 1.2)$ applied to 30\% of samples) and scale normalization. The model is trained on 100,000 web-collected human videos, with clips randomly sampled between 49 and 121 frames. To ensure a seamless pose transition from the reference image to the driving sequence, we mask the driving signals during the initial one-second segment of each training clip and prepend one second of zero frames (devoid of motion signals) at inference. Gemini 2.5~\cite{comanici2025gemini} serves as our (M)LLM for its superior multi-modal reasoning.

For end-to-end DreamActor-M2, we first synthesize large-scale data with diverse reference characters (humans,cartoons,animals) via the pre-trained pose-based model, followed by a two-stage filtering protocol. 
% For synthetic data, reference images include diverse categories (humans, cartoons, animals, etc.).
Specifically, an automated stage via Video-Bench~\cite{han2025video} filters for videos with average scores above 4.5, followed by manual verification to ensure rigorous motion consistency (between driving and generated videos) and identity preservation (between reference image and generated characters). This pipeline yields 60,000 high-quality video triplets for reliable end-to-end training. % Comprehensive details are provided in Appendix~\ref{sec: data_synthesis_more_details}.

% All experiments run on 16 NVIDIA A100 GPUs with 50,000 training steps and a batch size of 2. 
In our experiments, we train all stages for 50,000 training steps, with a batch size of 2.
The LoRA rank is set to 256. We optimize via AdamW with a learning rate of $5\times10^{-5}$ and weight decay of 0.01.

\textbf{Evaluation Metrics.}
Most evaluation metrics, such as FID-FVD~\citep{balaji2019fid-fvd}, FVD~\citep{unterthiner2018fvd}, and CD-FVD~\citep{ge2024content}, rely on comparisons with ground-truth videos, which are unavailable in cross-identity animation scenarios. As a result, these metrics fail to accurately reflect model performance. Moreover, prior studies have revealed that these metrics are often inconsistent with human judgment~\citep{huang2024vbench}. 
To address this, we adopt Video-Bench’s human-aligned automatic protocol~\cite{han2025video}, focusing on four key perceptual dimensions: \textit{Imaging Quality}, \textit{Motion Smoothness}, \textit{Temporal Consistency}, and \textit{Appearance Consistency}. All dimensions (automatic/human evaluation) use a 1–5 scale (1=very poor, 2=poor, 3=moderate, 4=good, 5=excellent), enabling comprehensive, reliable evaluation in real-world scenarios.

\begin{table*}[tb]
\caption{Quantitative comparisons with automatic evaluations and human evaluations on \texttt{\textit{AW}Bench}.}
\label{tab:comparison_awbench}
% \vspace{-8pt}
\centering
\setlength{\tabcolsep}{3pt}
\resizebox{\linewidth}{!}{
\begin{tabular}{l|cccc|ccc}
\toprule
\textbf{\multirow{3}{*}{Method}} 
  & \multicolumn{4}{c|}{\textbf{Automatic Evaluations (Video-Bench)}}  % 补充竖线，与列格式匹配
  & \multicolumn{3}{c}{\textbf{Human Evaluations}}\\ 
\cmidrule(lr){2-5} \cmidrule(lr){6-8}
& \textbf{\makecell{Imaging\\Quality $\uparrow$}} 
& \textbf{\makecell{Motion\\Smoothness $\uparrow$}}  
& \textbf{\makecell{Temporal\\Consistency $\uparrow$}}
& \textbf{\makecell{Appearance\\Consistency $\uparrow$}}
& \textbf{\makecell{Imaging\\Quality $\uparrow$}} 
& \textbf{\makecell{Motion\\Consistency $\uparrow$}}
& \textbf{\makecell{Appearance\\Consistency $\uparrow$}}\\ 
\midrule
Animate-X++~\cite{tan2025animate} 
  & 3.45 & 3.42 & 4.15 & 3.21 
  & 3.18 $\pm$ 0.23 & 2.95 $\pm$ 0.29 & 2.86 $\pm$ 0.34 \\ 
MTVCrafter~\cite{ding2025mtvcrafter} 
  & 3.71 & 3.81 & 4.02 & 3.53 
  & 3.35 $\pm$ 0.26 & 3.26 $\pm$ 0.28 & 3.07 $\pm$ 0.36 \\ 
DreamActor-M1~\cite{luo2025dreamactor}  
  & 4.17 & 3.92 & 4.21 & 4.06 
  & 3.96 $\pm$ 0.21 & 3.72 $\pm$ 0.26 & 3.54 $\pm$ 0.31 \\ 
Wan2.2-Animate~\cite{cheng2025wan} 
  & 4.05 & 4.06 & 4.17 & 3.92 
  & 3.91 $\pm$ 0.20 & 3.83 $\pm$ 0.25 & 3.51 $\pm$ 0.30 \\
\midrule
\textbf{Ours (Pose-based DreamActor-M2)} 
  & \textbf{4.68} & \textbf{4.53} & \textbf{4.61} & \textbf{4.28} 
  & \textbf{4.23 $\pm$ 0.19} & \textbf{4.18 $\pm$ 0.24} & \textbf{4.12 $\pm$ 0.28} \\
\textbf{Ours (End-to-End DreamActor-M2)} 
  & \textbf{4.72} & \textbf{4.56} & \textbf{4.69} & \textbf{4.35} 
  & \textbf{4.27 $\pm$ 0.18} & \textbf{4.24 $\pm$ 0.23} & \textbf{4.20 $\pm$ 0.29} \\
\bottomrule
\end{tabular}}
% \vspace{-6pt}
\end{table*}

\begin{figure}[tb]
    \centering
    \includegraphics[width=\columnwidth]{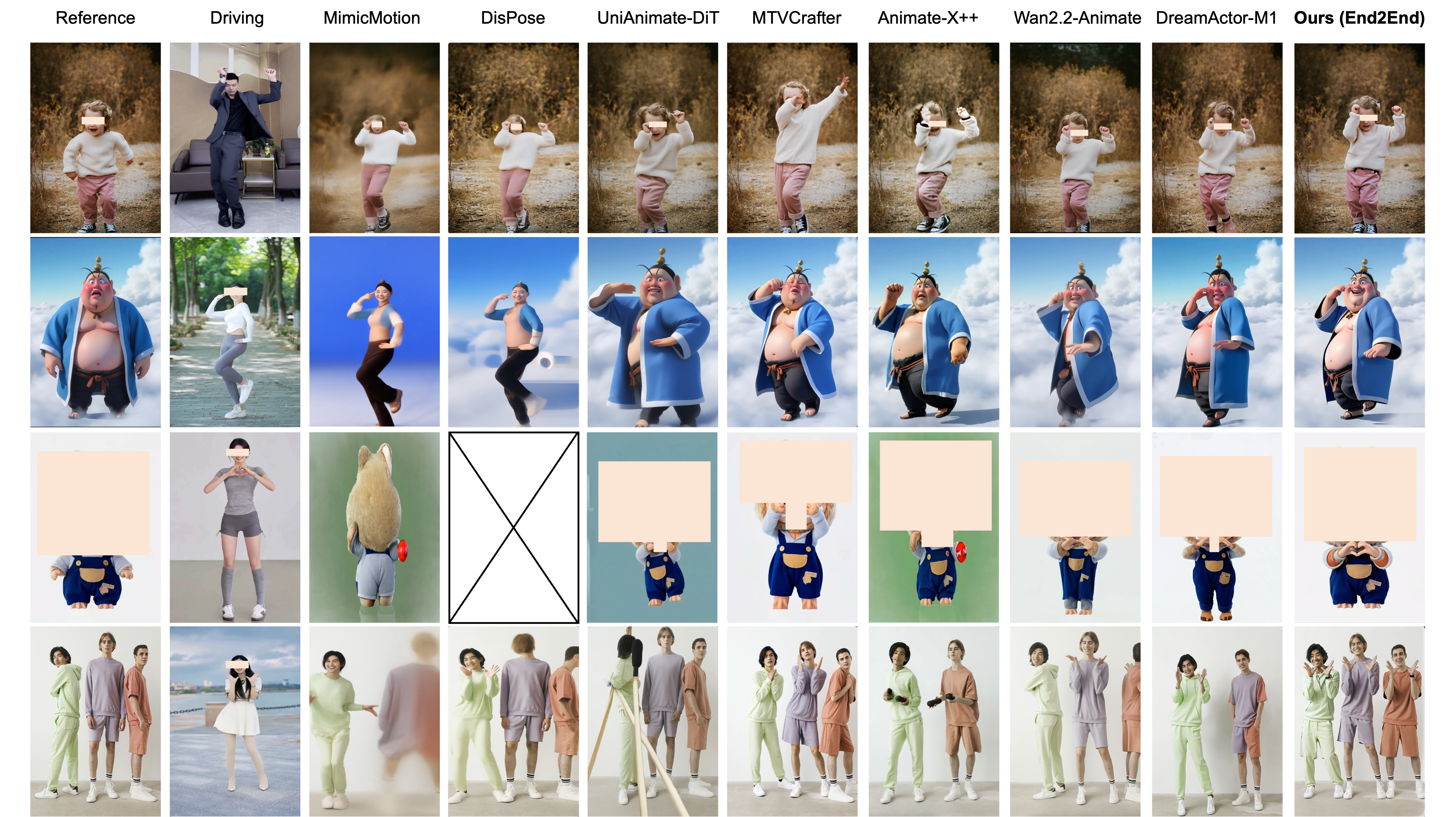}
    % \vspace{-0.4cm}
    \caption{Qualitative comparisons between our method and state-of-the-art approaches on \texttt{\textit{AW}Bench}.}
    \label{fig: sota_comparison}
\end{figure}

\subsection{Quantitative Comparison}
\textbf{Automatic evaluations and Human evaluations.}
We evaluate the model’s performance with \texttt{\textit{AW}Bench} to evaluate performance in more diverse scenarios. This test set comprises 60 human-to-human and 40 human-to-cartoon animation pairs. We compare our method with several recent state-of-the-art visual approaches. Due to the inherent lack of ground-truth in cross-identity tasks, we adopt Video-Bench~\cite{han2025video} for automatic evaluation, focusing on the four key metrics mentioned above. Furthermore, we conduct a user study involving 12 participants, who evaluated 100 randomly selected samples per method on a 5-point scale. As summarized in Tab.~\ref{tab:comparison_awbench}, our DreamActor-M2 variants outperform all competitors across all automatic metrics, achieving significant improvements in every evaluation dimension. 
This superiority is further mirrored in the human evaluation, 
\begin{wrapfigure}{r}{0.60\textwidth} %
  \includegraphics[width=\linewidth]{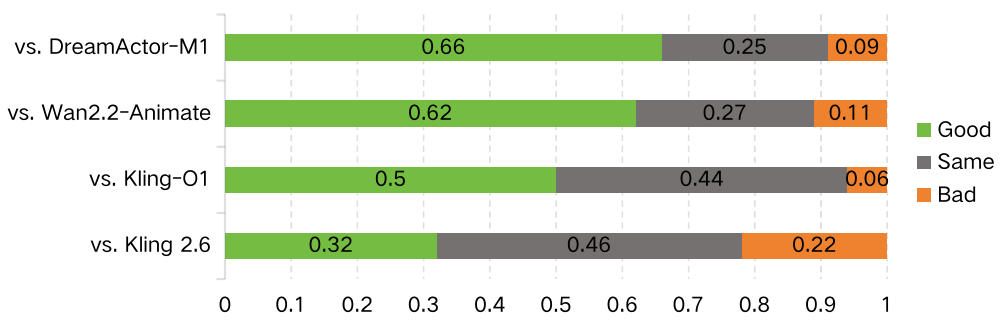}
    \caption{GSB comparison of DreamActor-M2 and other products.}
    \label{fig:gsb_comparison}
    % \vspace{-10pt} 
\end{wrapfigure}
\noindent where our methods surpass existing baselines by a substantial margin. The consistency between the automated metrics and human evaluation further underscores the superior generation quality and robustness of our DreamActor-M2 framework.

\textbf{GSB comparison with other products.} We conduct a GSB subjective evaluation to compare DreamActor-M2 with mainstream platform-level products based on a same testing dataset. The results in Fig.~\ref{fig:gsb_comparison} show DreamActor-M2 matches the overall performance of Kling 2.6 with a +9.66\% GSB lead, and outperforms other products by a significant margin: +43.66\% over Kling-O1, +51.43\% over Wan2.2-Animate, and +57.04\% over the prior DreamActor-M1. These results fully validate the improvements of DreamActor-M2, which delivers competitive and leading subjective performance against industry platform-level products.

\begin{figure}[tb]
    \centering
    \includegraphics[width=1\columnwidth]{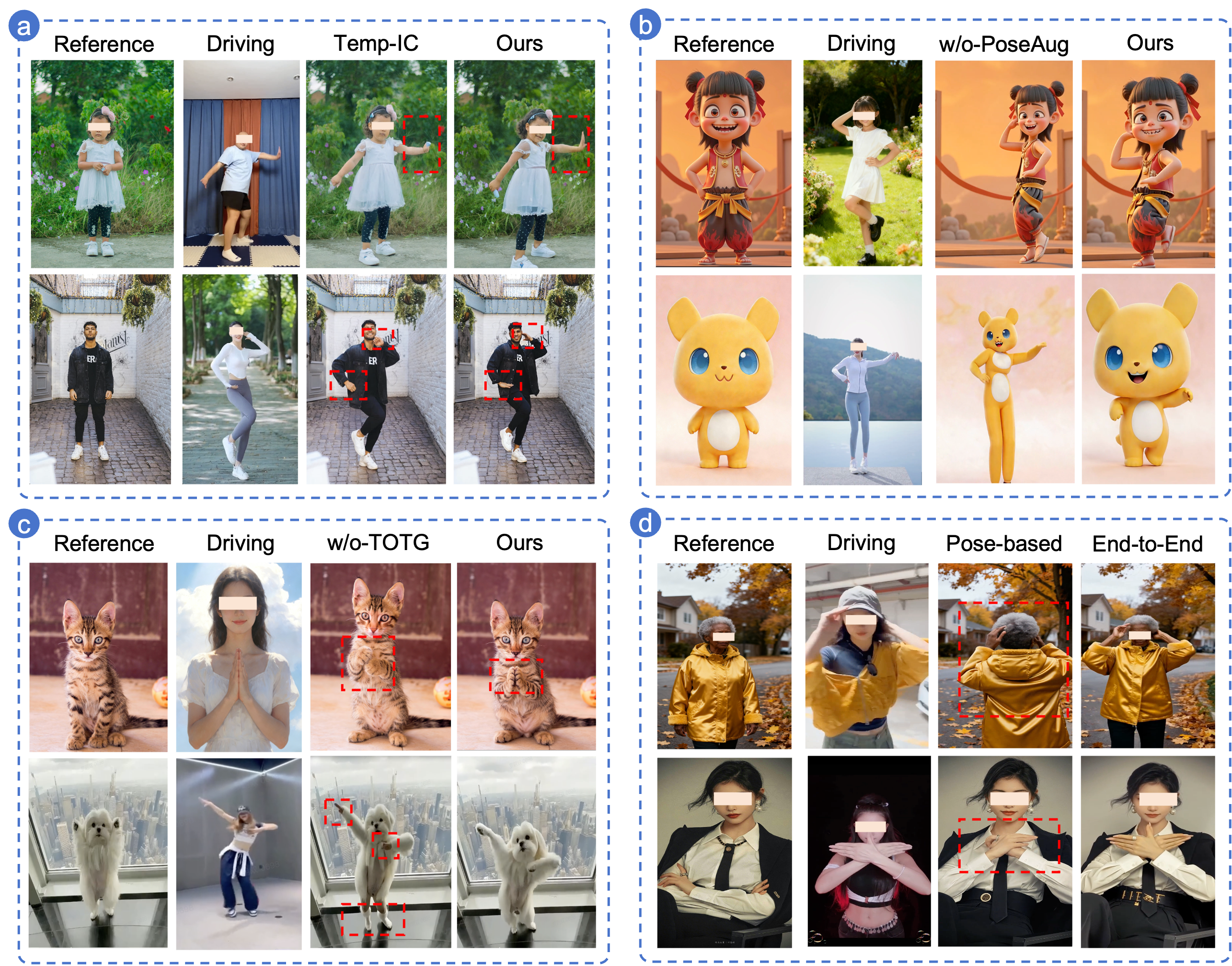}
    % \vspace{-0.8cm}
    \caption{Qualitative visualization for ablation study.}
    \label{fig: ablation_study}
\end{figure}

\subsection{Qualitative Results}
\textbf{Comparison with State-of-the-arts.} Fig.~\ref{fig: sota_comparison} presents a qualitative comparison across diverse animation scenarios, ranging from standard intra-domain driving to challenging cross-domain transfer and multi-person scenarios. In the first row, DreamActor-M2 demonstrates plausible visual fidelity, maintaining stringent identity preservation and motion alignment where baselines exhibit blurring. The second row highlights its remarkable body shape preservation and faithful motion alignment with driving inputs, outperforming others in capturing fine-grained motion details. In the third row, the accurate generation of the "heart gesture" validates the model’s superior grasp of motion semantics. Furthermore, the fourth row underscores the robustness of our method in one-to-many driving scenarios, a demanding task where competitors typically suffer from severe visual artifacts or structural collapse. Overall, these results demonstrate that DreamActor-M2 excels in handling heterogeneous animation tasks with exceptional identity preservation and motion consistency.

\begin{figure*}[tbp]
    \centering
    \includegraphics[width=0.9\columnwidth]{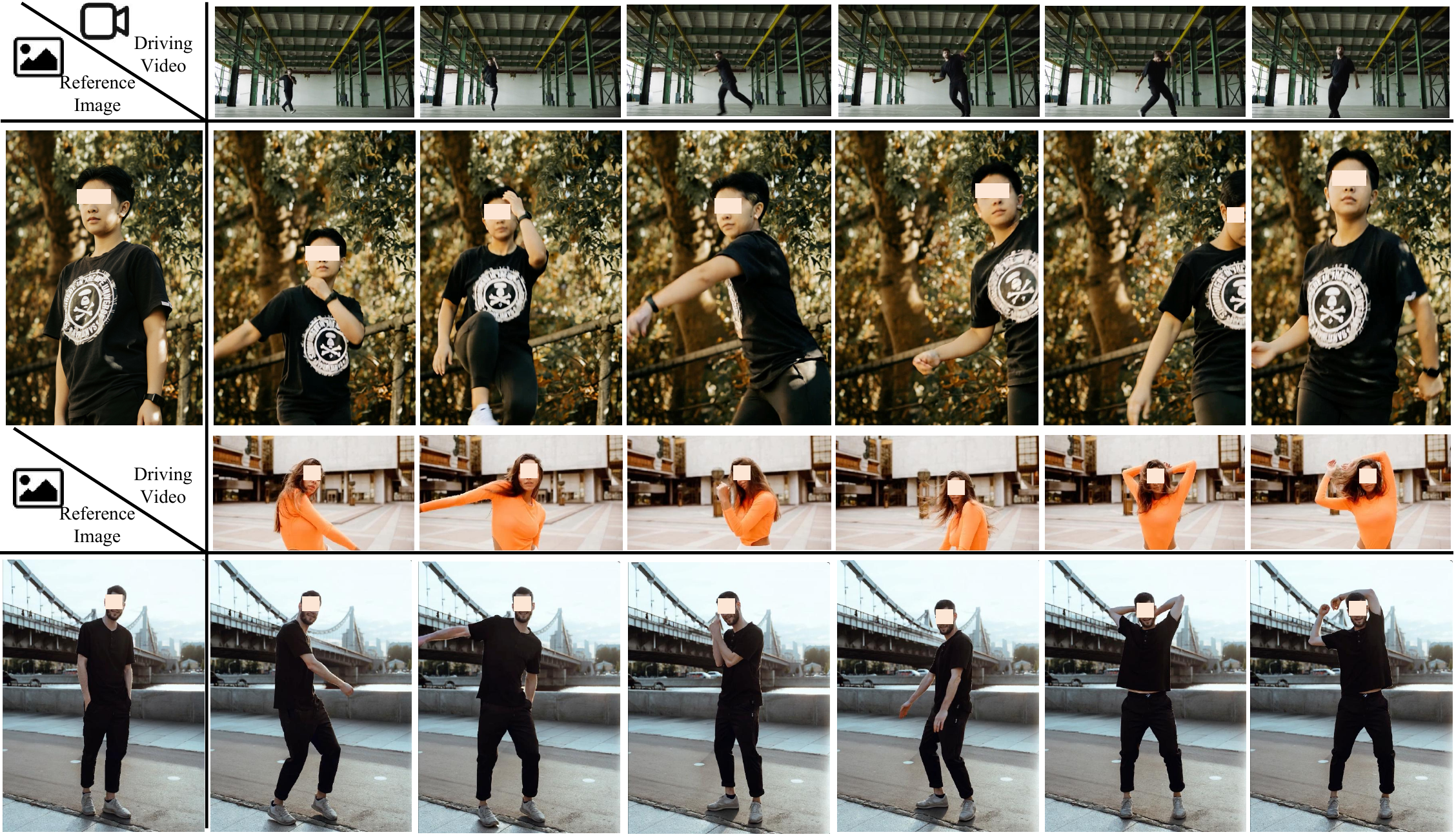}
    %\vspace{-15pt}
    % \vspace{-0.4cm}
    \caption{Qualitative visualization for various shot types.}
    \label{fig: different_frame_types_different_shot}
\end{figure*}

\begin{figure*}
    \centering
    % \vspace{-0.2cm}
    \includegraphics[width=0.9\columnwidth]{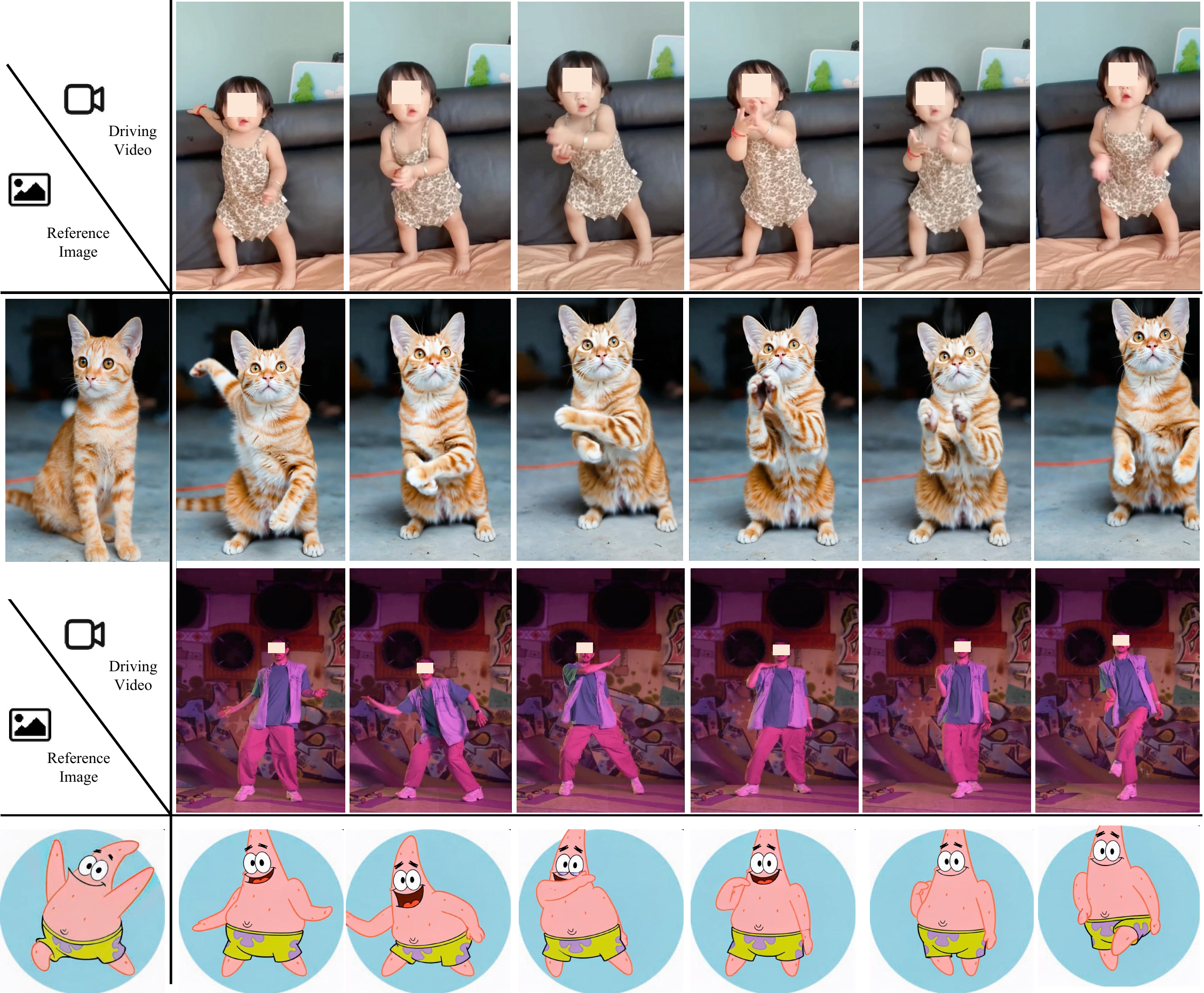}
    %\vspace{-15pt}
    % \vspace{-0.4cm}
    \caption{Qualitative visualization for reference character types.}
    \label{fig: different_frame_types_different_reference}
\end{figure*}

\begin{figure*}[tbp]
    \centering
    \includegraphics[width=0.9\columnwidth]{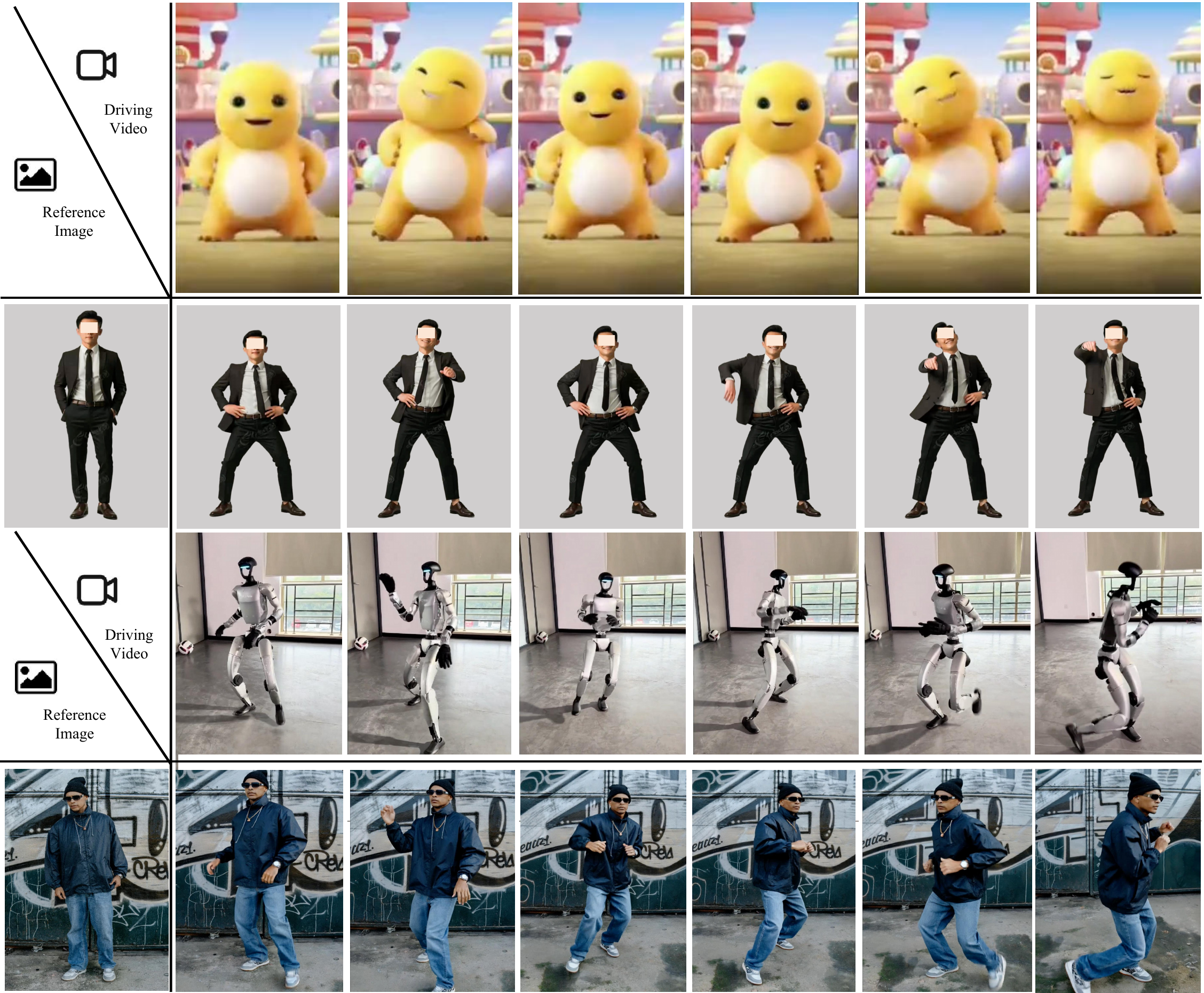}
    %\vspace{-15pt}
    % \vspace{-0.4cm}
    \caption{Qualitative visualization for various driving character types.}
    \label{fig: different_frame_types_different_driving}
\end{figure*}

\begin{figure*}
    \centering
    \includegraphics[width=0.9\columnwidth]{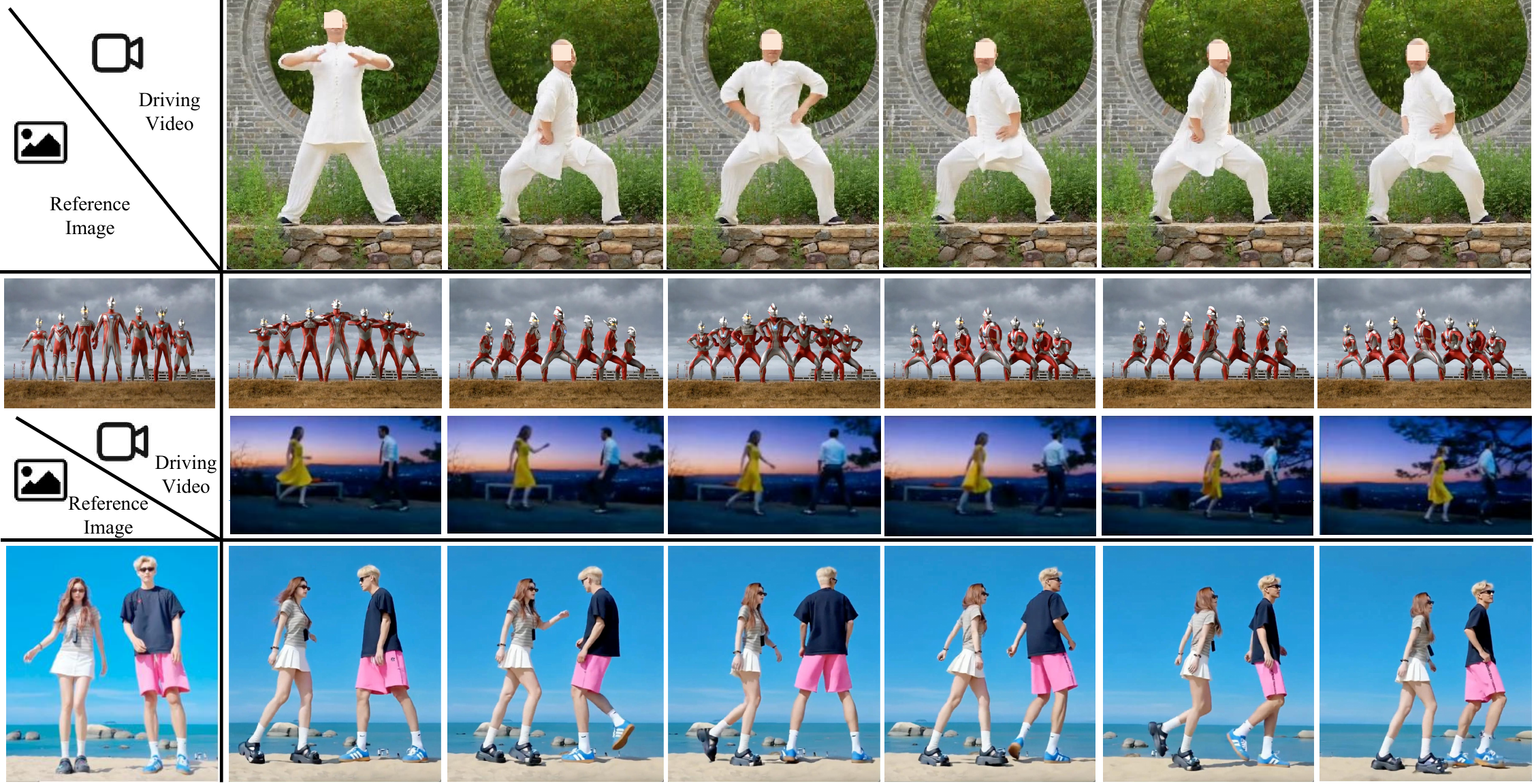}
    %\vspace{-15pt}
    % \vspace{-0.4cm}
    \caption{Qualitative visualization for multi-person settings.}
    \label{fig: different_frame_types_multi_persons}
\end{figure*}

\textbf{Visualization for generalization.}
To qualitatively evaluate DreamActor-M2's generalization, we conduct experiments across four key scenarios as shown in Fig.~\ref{fig:intro}.
\textbf{(1) Shot types}: The model demonstrates exceptional cross-morphology mapping. Notably, in \textit{Half2Full} tasks, despite the absence of lower-body driving signals, DreamActor-M2 leverages the pre-trained backbone’s generative priors to synthesize plausible lower-body motions while maintaining precise upper-body synchronization.
\textbf{(2) Reference characters}: Moving beyond human-centric constraints, our model supports diverse subjects. Fig.~\ref{fig:intro} showcases high-fidelity animation for animals (rabbit), objects (juice bottle), and cartoon characters (Detective Conan, Pikachu).
\textbf{(3) Driving characters}: Our end-to-end framework generalizes to non-human motion sources. Fig.~\ref{fig:intro} demonstrates its capacity to process diverse driving signals (e.g., \textit{Animal2Animal} and \textit{Cartoon2Cartoon}), enabling transfers that transcend traditional human motion priors.
\textbf{(4) Multi-person scenarios}: DreamActor-M2 excels in both \textit{One2Multi} and \textit{Multi2Multi} tasks. It effectively synchronizes motion across multiple distinct characters or maps complex multi-person dynamics to new groups without structural collapse.
More qualitative results are provided in Fig.~\ref{fig: different_frame_types_different_shot}, Fig.~\ref{fig: different_frame_types_different_reference}, Fig.~\ref{fig: different_frame_types_different_driving} and Fig.~\ref{fig: different_frame_types_multi_persons}.

\subsection{Ablation Study}
\begin{wraptable}{r}{0.6\textwidth} 
% \vspace{-10pt} % 压缩顶部间距（避免表格被顶出可视区域）
\centering
\small % 缩小字体（适配 0.6\textwidth 宽度，避免溢出）
\setlength{\tabcolsep}{4pt} % 调整列间距（避免表格过宽）
\begin{tabular}{l|ccc}
\toprule % 替换 multirow 兼容的横线（避免排版冲突）
\multirow{2}{*}{\textbf{Method}}  & \multicolumn{3}{c}{\textbf{Human Evaluations}} \\ 
\cmidrule(lr){2-4}
& \textbf{\makecell{Imaging\\Quality}} $\uparrow$ 
& \textbf{\makecell{Motion\\Consistency}} $\uparrow$
& \textbf{\makecell{Appearance\\Consistency}} $\uparrow$ \\ 
\midrule
Temp-IC          & 4.12  & 3.98 & 4.06 \\
w/o-PoseAug      & 4.15  & 3.80 & 3.92 \\
w/o-TOTG         & 4.21  & 3.85 & 4.08 \\
\midrule 
\textbf{Ours (Pose-based)} & \textbf{4.23} & \textbf{4.18} & \textbf{4.12}  \\
\bottomrule
\end{tabular} 
% \vspace{-5pt} % 压缩表格与标题间距
\caption{Ablation study on proposed DreamActor-M2 framework.}
\label{tab:ablation}
% \vspace{-8pt} % 压缩标题底部间距（避免影响后续排版）
\end{wraptable}

We conduct ablation studies to evaluate each core component of DreamActor-M2. First, we compare our spatiotemporal injection approach against the temporal injection method (Temp-IC). As shown in Tab.~\ref{tab:ablation}, DreamActor-M2 yields better overall generation quality. Moreover, as illustrated in Fig.~\ref{fig: ablation_study} (a), it better preserves intricate structural details—such as hand gestures—underscoring its advantages in spatial fidelity. 
Next, we ablate pose augmentation (w/o-PoseAug) to evaluate its necessity. As shown in Tab.~\ref{tab:ablation} and Fig.~\ref{fig: ablation_study}(b), pose augmentation consistently improves generation quality and is crucial for maintaining the reference subject's original body shape.
Finally, comparing DreamActor-M2 with a variant lacking target-oriented text guidance (w/o-TOTG) validates the impact of LLM-driven refinement. 
As demonstrated in Tab.~\ref{tab:ablation} and Fig.~\ref{fig: ablation_study} (c), injecting target-oriented text information via the LLM yields superior results, enabling DreamActor-M2 to better reconstruct semantically specific motions and preserve character identity. Fig.~\ref{fig: ablation_study} (d) shows our end-to-end model outperforms the pose-based counterpart in challenging 2D keypoint detection scenarios, including direction ambiguity and hand overlapping.
%\section{Conclusion and Future Work}
\section{Conclusion}
We introduce DreamActor-M2, a universal framework for character animation. At its core is a spatiotemporal in-context learning strategy that integrates motion and reference signals into a unified representation. 
This design not only harnesses the pre-trained backbone's generative priors, but further facilitates an evolution toward direct end-to-end motion transfer from raw videos, thereby eliminating the need for explicit pose estimation.
This versatility enables application to increasingly diverse and challenging scenarios.
Extensive experiments, including benchmarks on our newly curated \texttt{\textit{AW}Bench}, demonstrate that DreamActor-M2 establishes a robust and unified paradigm with exceptional fidelity and generalization.

\section{Limitations and Future Works}
While DreamActor-M2 demonstrates robust performance across various scenarios, it occasionally struggles with complex interactions, such as two characters rotating around each other. This is primarily attributed to the scarcity of training data featuring motion trajectory crossing. In the future, we plan to curate more diverse datasets with intricate multi-person interactions to further extend the model's applicability.

\section{Ethics considerations}
In our data and experiments, a number of human-related images and videos are involved. Meanwhile, our framework is capable of implementing human image animation. Human image animation has possible social risks, like being misused to make fake videos.
The proposed technology could be used to create fake
videos of people, but existing detection tools can
spot these fakes. To reduce these risks, clear ethical rules
and responsible usage guidelines are necessary. We will
strictly restrict access to our core models and codes to prevent misuse. Images and videos are all from publicly available sources. If there are any concerns, please contact us
and we will delete it in time.

\newpage

\clearpage\cleardoublepage
\bibliographystyle{plainnat}
\bibliography{main}

% \clearpage

% \beginappendix

% \input{sections/appendix}

\end{document}